# Fuzzy Soft Set Based Classification for Gene Expression Data

Kalaiselvi.N, Hannah Inbarani.H

**Abstract** — Classification is one of the major issues in Data Mining Research fields. The classification problems in medical area often classify medical dataset based on the result of medical diagnosis or description of medical treatment by the medical practitioner. This research work discusses the classification process of Gene Expression data for three different cancers which are breast cancer, lung cancer and leukemia cancer with two classes which are cancerous stage and non cancerous stage. We have applied a fuzzy soft set similarity based classifier to enhance the accuracy to predict the stages among cancer genes and the informative genes are selected by using Entopy filtering.

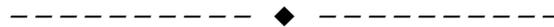

## 1 INTRODUCTION

Gene expression represents the activation level of each gene within an organism at a particular point of time. The classification is the process of predicting the classes among the huge amount of dataset by using some machine learning algorithms. The classification of different tumor types in gene expression data is of great importance in cancer diagnosis and drug discovery but it is more complex because of its huge size. DNA micro array technologies have made it possible to measure the expression levels of thousand of genes in a single experiment. There are multiple techniques available to analyze a gene expression profiles. A common characteristic of these techniques is selecting a subset of genes which is very informative for classification process and to reduce the dimensionality problem of profiles. In this research work ,Entropy based filtering approach is used for ranking the cancer genes and the top most genes were selected [1].Genes selected are classified by using Fuzzy Soft Sets and this classification approach reduced the complexity and increased the accuracy of classification compared with standard Fuzzy KNN algorithm and KNN. This paper also illustrates the effectiveness of the proposed approach over the other popular classification approaches such as K-Nearest Neighbor approach and Fuzzy KNN approach.

## 2 RELATED WORK

Soft set theory was initiated by Molodstov to solve some uncertainties among data which is not solved by traditional mathematical tools. He has shown several applications of this theory by solving many practical problems in various aspects like engineering, economics, and medical science etc [2]. Maji et al. have further studied the theory of soft sets and used this theory for some decision making problems[3][4]. They have also introduced the concept of fuzzy soft set theory for decision making problems. Aktas and cagman have introduced the notion of soft groups[5]. P.Majumdar et al. [6] have introduced generalized fuzzy soft set and they have studied the similarity measures of fuzzy soft set for medical diagnosis to detect the pneumonia disease among the ill persons[6]. Recently Milind M.Mushrif,et al have introduced a novel method for classification of natural textures using fuzzy soft set based classifier. Their new approach provides high accuracy for texture classification[7]. Bana Handaga et al. proposed numerical data classification for seven types of medical data and they provided higher accuracy [8]. Saberi et al. proposed gene selection method for cancer classification with less complexity[9].

## 3 RESEARCH MOTIVATION

The main challenges in gene expression data classification is its huge size and its vagueness.Many researchers are trying to achive the higher accuracy for gene classification by reducing or selecting the informative genes among the thousands of genes. For cancer classification, researchers have used some machine learning algorithms like Support Vector Machines, Principle Component Analysis etc., Rough Set Theory and Fuzzy logic are only used for Gene selection. In this paper we use soft set based classification method for cancer classification. In this algorithm genes are selected for dimensionality reduction by entropy filter approach and the genes are fuzzified based on its expression level and fuzzy soft set based classification method is applied for cancer classification.

## 4 METHODOLOGY

The methodology adopted in this work is given in Figure 1. In the first step, genes are preprocessed. In the second step , informative genes are selected by using entropy filtering. Then based on its expression value, genes are fuzzified. In the fourth step , fuzzy soft set based classification method is applied for gene classification.

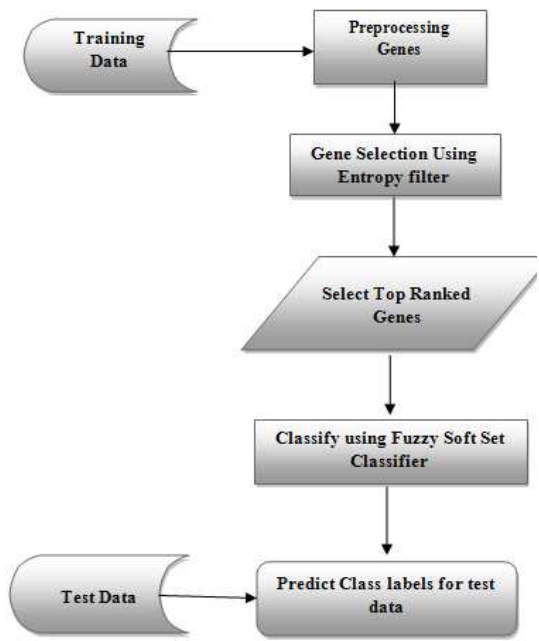

Figure 1: Methodology for classification of Gene Expression Data

## 5 PRELIMINARIES

### 5.1 Soft set theory

In this section, we describe the basic notions of soft sets, fuzzy soft sets and similarity measures between the fuzzy soft sets. Let U be initial universe of objects and E be a set of parameters in relation to objects in U. Parameters are often attributes, characteristics or properties of objects[2].

*Definition:* A pair (F, A) is called a soft set over U, where F is a mapping given by

$$F : A \rightarrow P(U).$$

In other words, a soft set over U is a parameterized family of subsets of the universe U. For $\varepsilon \in A$, $F(\varepsilon)$ may be considered as the set of $\varepsilon$-elements of the soft set (F,A) or as the set of $\varepsilon$-approximate elements of the soft set[2].

### 5.2 Fuzzy Soft Set Theory

Let $U$ be an initial universal set and let $E$ be set of parameters. Let $\tilde{P}(U)$ denote the power set of all fuzzy subsets of U. Let $A \subseteq E$ [3].

**Definition:** A pair ($\tilde{F}$,E) is called a fuzzy soft set over U, where $\tilde{F}$ is a mapping given by

$$\tilde{F}:A \rightarrow \tilde{P}(U).$$

In the above definition, fuzzy subsets in the universe U are used as substitutes for the crisp subsets of U. Hence it is easy to see that every (classical) soft set may be considered as a fuzzy soft set. Generally speaking $\tilde{F}(\varepsilon)$ is a fuzzy subset in U and it is called the fuzzy approximate value set of the parameter $\varepsilon$.

## 6 FUZZY SOFT SET BASED CLASSIFICATION FOR GENE EXPRESSION DATA

Given a Gene Expression dataset with m samples belonging to k known classes and n genes. where {$(g_1, c_1), (g_2, c_2), \ldots , (g_n, c_n)$}, where G = $g_1, g_2, \ldots g_n$ are the genes in the dataset k =$c_1, c_2, \ldots c_n$ are the classes in the dataset.

### 6.1 Gene selection

Gene selection is the process of selecting informative genes among huge number of genes. This process will reduce the complexity and dimensionality of the dataset. In this gene selection method correlation between the genes are calculated by using entropy based filter approach. For each and every gene, entropy value will be calculated and the low entropy valued genes are removed [1].

### 6.2 Entropy Filtering

The effectiveness of the genes was computed by using entropy filtering method. Entropy measures the uncertainty of random variables. For the measurement of interdependency of two random genes X and Y, Shannon's information theory is used [1].

$$H(X) = -\sum_i P(X_i) \log P(X_i) \quad (1)$$

H(X) is entropy value of individual gene X. By using Entropy filtering, Information Gain is computed for the random genes and depending upon the gain value, the genes may be removed or selected.

$$IG(X,Y)=H(X)+H(Y) – H(X,Y) \quad (2)$$

Figure 2 represents cluster dendogram analysis of Lung cancer genes before filtering and after filtering. Figure 3 represents the cluster dendogram analysis of Leukemia genes before and after filtering. Figure 4 represents the cluster dendogram analysis of Breast cancer genes before and after filtering.

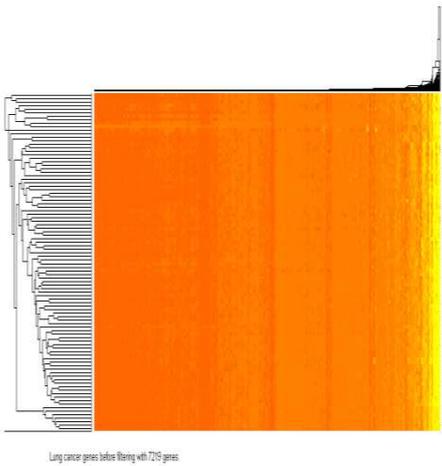

Figure 2(a): Cluster dendogram analysis of Lung Cancer Genes before Filtering – 7219 genes

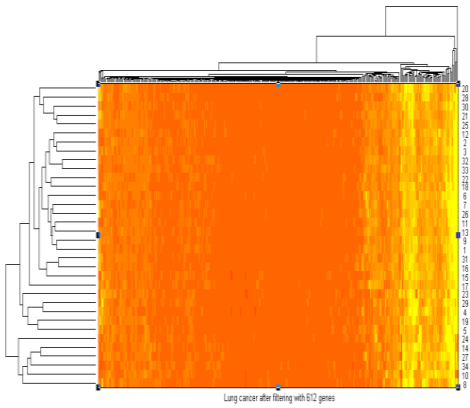

Figure 2(b): Cluster dendogram analysis of Lung Cancer Genes after filtering -612 genes

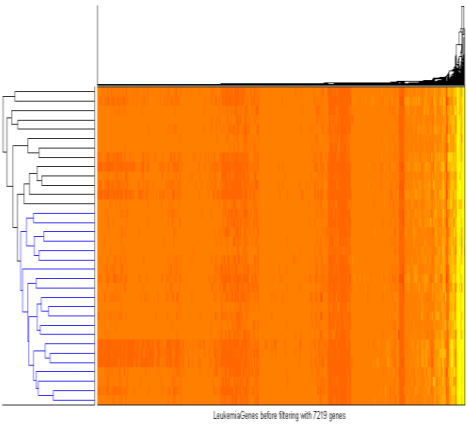

Figure 3(a):Cluster dendogram of leukemia genes before filtering - 7219 genes

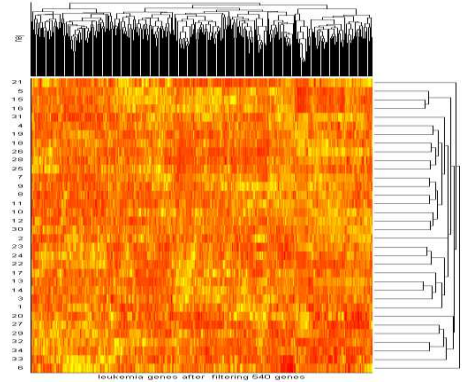

Fig 3(b)Cluster dendogram analysis of Leukemia Genes after filtering – 520 genes

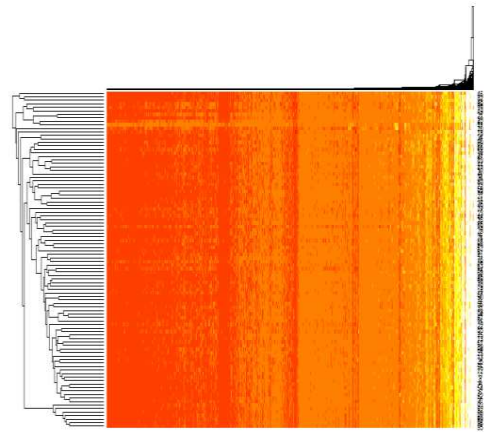

Fig.4(a) Cluster dendogram analysis of Breast cancer genes before filtering – 5400 genes

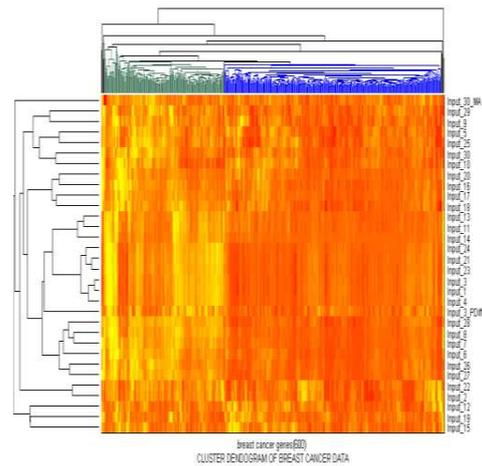

Fig.4(b) Cluster dendogram analysis of breast cancer genes after filtering - 600 genes

## 7 FUZZY SOFT SET BASED CLASSIFICATION

This approach is used to classify numerical data based on the theory of fuzzy soft sets by Bana Handaga and Mustafa Mat Deris[8]. They applied this approach for medical datasets with higher accuracy. In this paper, fuzzy soft set similarity algorithm is applied for three cancer gene expression datasets namely Lukemia, Lung cancer and Breast cancer[10],[11].

### 7.1 ALGORITHM

**Pre-processing phase**

1. Fuzzify feature vector vaues uisng S- shaped membership function and Z- shaped membership function $E_{ci}$, I = 1,2, …, N for all genes, Training dataset and test dataset.

$C_i$ represshts classes in the dataset such as benign and malignant.

2. Select informative genes by using Entropy filter approach using (1). The genes with low entropy values are removed.

**Training phase**

1. Give N Genes as input along with class values.
2. Calculate the cluster center vector $E_{ci}$ using (3). $E_{ci}$=1/N

$$\sum_{i=1}^{N} E_{ci} \quad (3)$$

3. Obtain a fuzzy set model for class $C$, ($\bar{F}_{ci}$, E), is a cluster center for class C having N features(samples).
4. Repeat the process for all C classes.

**Classification phase**

1. Get the unknown class data.
2. Obtain a fuzzy soft sets model for unknown class data, ($\tilde{G}$, E) and ($\bar{F}_c$, C) for each C using (4)
3. Calculate similarity between ($\tilde{G}$, E) and ($\bar{F}_c$, E) for each c classes by using (5)

Similarity between two fuzzy soft set was measured by using the following formula.

$$S(F,G) = 1 - \frac{\sum_{j=1}^{n}|\tilde{F}_{ij} - \tilde{G}_{ij}|}{\sum_{j=1}^{n}|\tilde{F}_{ij} + \tilde{G}_{ij}|} \quad (4)$$

$\tilde{F}_{ij}$(Fuzzv soft set of known class data)= $\mu_{\bar{F}}(x_i)$.
$\tilde{G}_{ij}$(Fuzzy softset of unknown class data)= $\mu_{\bar{G}}(x_i)$.

4. Assign the unknown data to class C if similarity is maximum.

$$C = \arg[\max_{c=1}^{W} S(\tilde{G}, \tilde{F}_c)] \quad (5)$$

## 8 EXPERIMENTAL RESULTS

### 8.1 Dataset

The Data set is collected from public microarray data repository [10], [11]. In this work, the three cancer gene expression datasets are taken namely lung cancer dataset with 7219 genes and 64 samples, Leukemia dataset with 7219 genes and 92 samples and Breast cancer dataset with 5400 genes and 34 samples.

A comparative analysis of Fuzzy Soft Set based classification is made with benchmark algorithms such as Fuzzy K Nearest Neighbor (KNN) algorithm and K Nearest Neighbor classification based on classification accuracy measures[8].

### 8.2 K Nearest Neighbor Algorithm

K-nearest neighbor is a supervised learning algorithm where the result of new instance query is classified based on majority of K-nearest neighbor category. The purpose of this algorithm is to classify a new object based on attributes and training samples. This algorithm used neighborhood classification as the prediction value of the new query instance[12].

1. Determine parameter K = number of nearest neighbors
2. Calculate the distance between the query-instance and all the training samples
3. Sort the distance and determine nearest neighbors based on the K-th minimum distance
4. Gather the category Y of the nearest neighbors.
5. Use simple majority of the category of nearest neighbors as the prediction value of the query instance

### 8.3 Fuzzy K Nearest Neighbor Algorithm

The fuzzy K-nearest neighbor algorithm assigns class membership to a sample vector rather than assign vector to a particular class. The advantage is that no arbitrary assignments are made by the algorithm. In addition, the vector membership values should provide a level of assurance to accompany the resultant classification.

The basis of algorithm is to assign membership as a function of the vector's distance from its K-nearest neighbors and those neighbor's memberships in the possible classes. The fuzzy algorithm is similar to the crisp version in the sense that it must also search the labeled sample set for the K-nearest neighbor[12].

In this section our Fuzzy Soft Set Gene Classification is com-

pared with KNN and Fuzzy KNN based on some validation measures. The comparative analysis shows best performance of Fuzzy Soft Set based Gene Classification approach than KNN and Fuzzy KNN.

## 8. 4 Validation measures:

**Precision**

Precision is a measure of the accuracy provided that a specific class has been predicted[8]. It is defined by:

**Precision=tp / (tp+fp)**

where tp and fp are the numbers of true positive and false positive predictions for the considered class

**Recall**

Recall is a measure of the ability of a prediction model to select instances of a certain class from a data set. It is commonly also called sensitivity, and corresponds to the true positive rate[8].

**Recall /Sensitivity = tp / (tp + fn)**

**Specificity**

Recall/sensitivity is related to specificity, which is a measure that is commonly used in two class problems where one is more interested in a particular class. Specificity corresponds to the true-negative rate.

**Specificity = tn / (tn + fp)**

*Overall classification Accuracy*

Accuracy is the overall correctness of the model and is calculated as the sum of correct classifications divided by the total number of classifications [8].

**Accuracy = (True Classification)/(Total no of cases)**

Table 1,2 and 3 represents performance analysis of Classification algorithms on lung cancer dataset, Leukemia and breast cancer dataset. Figures 5,6 and 7 show the comparative analysis of classification algorithms for gene expression datasets. Figure 8 shows the overall accuracy of classification algorithms.

Table 1: Performance Analysis on Lung cancer Gene Expression Data

| Accuracy Measures | Classification algorithms | | |
|---|---|---|---|
| | Fuzzysoft set based classification | K-nearest neighbor Approach | Fuzzy KNN |
| Precision | 0.78 | 0.71 | 0.79 |
| Sensitivity | 0.75 | 0.75 | 0.77 |
| Specificity | 0.86 | 0.79 | 0.82 |

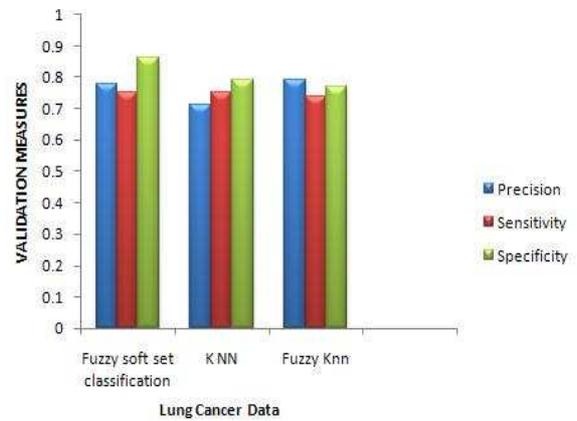

Figure 5: Comparative analysis of classification algorithms for Lung Cancer dataset

Table 2: Performance Analysis of Classification approaches for Leukemia cancer Gene Expression Data

| Accuracy Measures | Classification algorithms | | |
|---|---|---|---|
| | Fuzzy Soft Set Based Classification | K-Nearest Neighbor Approach | Fuzzy KNN Algorithm |
| Precision | 0.79 | 0.65 | 0.65 |
| Sensitivity | 0.85 | 0.85 | 0.82 |
| Specificity | 0.84 | 0.75 | 0.76 |

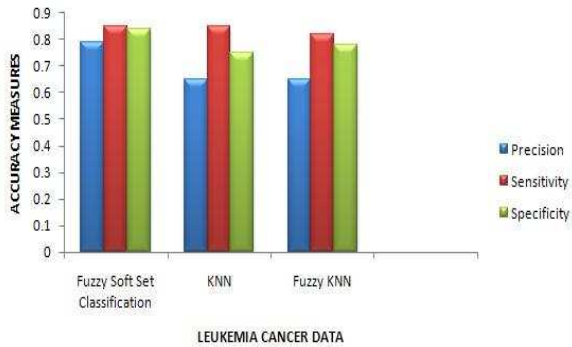

Fig.6 Comparative analysis of classification algorithms for Leukemia dataset

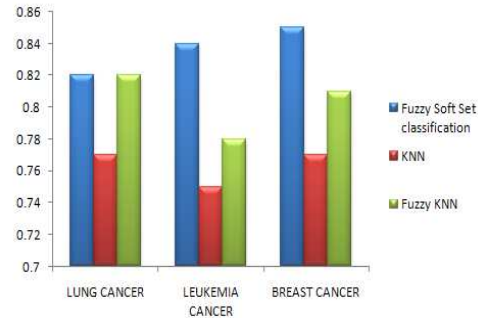

Fig.8 Overall accuracy of Classification algorithms

## 9. CONCLUSION

Classification is the most important technique in microarray technology. This technique is used for prediction of classes among the genes or samples. Prediction plays an important role in biomedical field for disease stage prediction and drug discovery. In this paper, classification technique is applied for predicting the cancer types among the genes in various cancer gene expression datasets such as leukemia cancer gene expression dataset with two classes tumor and non tumor, lung cancer gene expression dataset with two classes tumor and normal and Breast cancer gene expression dataset with two classes benign and malignant.

In this work, Fuzzy Soft Set Gene Classification is proposed for classification of Gene expression data. The classification accuracy of the proposed approach is compared with the KNN and Fuzzy KNN Algorithm. The experimental analysis illustrates the effectiveness of fuzzy soft set approach over the other two approaches. In future, it can be applied for other data sets also.

Table 3: Performance Analysis of classification approaches for Breast cancer Gene Expression Data

| Accuracy Measures | Classification algorithms | | |
|---|---|---|---|
| | Fuzzy Soft Set Based Classifiction | K-nearest Neighbor Approach | Fuzzy KNN Algorithm |
| Precision | 0.79 | 0.65 | 0.65 |
| Sensitivity | 0.85 | 0.85 | 0.82 |
| Specificity | 0.84 | 0.75 | 0.76 |

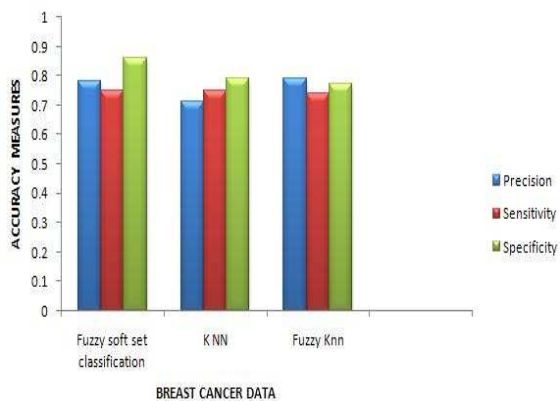

Fig .7 Comparative analysis of classification algorithms for Breast cancer dataset